\documentclass{ecai}
\usepackage{times}
\usepackage{graphicx}
\usepackage{latexsym}
\usepackage[numbers]{natbib}
 \usepackage{amsmath,amssymb,amsfonts, amsthm}

\usepackage{textcomp}

\usepackage{scalerel}
\usepackage{booktabs}
\usepackage{algorithmic}
\usepackage[ruled,norelsize]{algorithm2e}
\usepackage{tabularx}
\usepackage{subcaption}
\usepackage{xspace}

\theoremstyle{definition}
\newtheorem{definition}{Definition}

\newcommand{\model}{Temporal Progressive Growing Sampling\xspace}
\newcommand{\modelshort}{TPG\xspace}

\title{Bridging the Gap Between Training and Inference for Spatio-Temporal Forecasting}

\author{Hong-Bin Liu \and Ickjai Lee \institute{College of Science and Engineering, James Cook University,
Australia. email: $\{$hongbin.liu, ickjai.lee$\}$@jcu.edu.au} }

\begin{document}

\maketitle
\begin{abstract}
Spatio-temporal sequence forecasting is one of the fundamental tasks in spatio-temporal data mining. It facilitates many real world applications such as precipitation nowcasting, citywide crowd flow prediction and air pollution forecasting. Recently, a few Seq2Seq based approaches have been proposed, but one of the drawbacks of Seq2Seq models is that, small errors can accumulate quickly along the generated sequence at the inference stage due to the different distributions of training and inference phase. That is because Seq2Seq models minimise single step errors only during training, however the entire sequence has to be generated during the inference phase which generates a discrepancy between training and inference. In this work, we propose a novel curriculum learning based strategy named \model to effectively bridge the gap between training and inference for spatio-temporal sequence forecasting, by transforming the training process from a fully-supervised manner which utilises all available previous ground-truth values to a less-supervised manner which replaces some of the ground-truth context with generated predictions. To do that we sample the target sequence from midway outputs from intermediate models trained with bigger timescales through a carefully designed decaying strategy. Experimental results demonstrate that our proposed method better models long term dependencies and outperforms baseline approaches on two competitive datasets.

\end{abstract}

\section{Introduction}

Spatio-Temporal Sequence Forecasting (STSF) is one of the fundamental tasks in spatio-temporal data mining~\cite{2018arXiv180806865S}. It facilitates many real world applications such as precipitation nowcasting \cite{Shi:2015:CLN:2969239.2969329}, citywide crowd flow prediction \cite{ZHANG2018147,8594916} and air pollution forecasting \cite{Yi:2018:DDF:3219819.3219822}. It is to predict future values based on a series of past observations. STSF is formally defined as below:
\theoremstyle{definition}
\begin{definition}%{\textit{Spatio-Temporal Sequence Forecasting.}}
	Given a length-$T$ matrix sequence $\mathcal{S}=\left[\mathcal{X}_{1}, \mathcal{X}_{2}, \ldots, \mathcal{X}_{T}\right]$. Each matrix $\mathcal{X}_{t} \in \mathcal{S}$ consists of measurements of coordinates at time-step $t$. STSF is to predict a sequence of corresponding measurements of following $k$ time-steps based on the past observations of $\mathcal{S}$, denoted as $\hat{\mathcal{P}} = \left[\hat{\mathcal{X}}_{T+1}, \hat{\mathcal{X}}_{T+2}, \ldots, \hat{\mathcal{X}}_{T+k}\right]$.
\end{definition}

With the recent advances of Seq2Seq model \cite{NIPS2014_5346} in sequence modelling such as Neural Machine Translation (NMT) and speech recognition, researchers adapt Seq2Seq to model STSF as sequence modelling. In particular, both DCRNN \cite{li2018diffusion} and PredRNN \cite{NIPS2017_6689} utilised an RNN-based encoder to encode a source sequence $\mathcal{S}$ into a feature matrix, such feature matrix will then be decoded recursively conditioned on previous contexts into the target sequence $\hat{\mathcal{P}}$ with a separated RNN-based decoder. During the training phase, the previous contexts become ground-truth observations. However, during the inference phase, the previous contexts are drawn from the model itself as no ground-truth observations are available for the decoder. The cause of the discrepancy between the training and inference stages is so called \textit{exposure bias} \cite{RanzatoCAZ15}. Resulting small errors caused by the bias are quickly accumulated to become a large error along the generated sequence at the inference stage. One intuitive solution is to unify the training and inference phases by using previously generated contexts instead of ground-truth values during training. However, this causes the model more difficult or even unable to converge \cite{Bengio:2015:SSS:2969239.2969370,Venkatraman:2015:IMP:2888116.2888137}. Bengio \textit{et al.} introduced \textit{Scheduled Sampling} \cite{Bengio:2015:SSS:2969239.2969370} to overcome such problem by gradually transform between these two strategies which shows significant improvements and has been widely used in NMT systems. DCRNN and PredRNN adapt Scheduled Sampling into STSF which shows some improvements. However, we argue that simply adopting Scheduled Sampling from NMT to STSF is not ideal even though they both are for sequence modelling due to two clear distinctions. First, NMT system is basically for the word level classification problem that optimises the cross entropy loss conditioned on  the source sequence and previous contexts, whereas STSF is for the regression problem that optimises a regression loss such as Mean Square Error (MSE) and Mean Absolute Error (MAE). Second, two consecutive words in NMT systems are semantically close to each other, and small errors caused by the training bias could result in a completely different translation. However, two measurements in two consecutive time steps in STSF are geographically close and contiguous to each other, errors make by previous steps could result a confusion to the local trend of prediction that leads to bigger gap to the long term predictions.

Motivated by the above observations, we bridge the gap between training and inference for spatio-temporal forecasting by introducing a novel coarse-to-fine hierarchical sampling method named \emph{{\large \textbf{T}}emporal {\large \textbf{P}}rogressive {\large \textbf{G}}rowing} Sampling, in short \modelshort.  The idea is to transform the training process from a fully-supervised manner which utilises all available previous ground-truth to a less-supervised manner which replaces some of the ground-truth contexts with generated predictions. To do that we also sample the target sequence from midway outputs from intermediate models trained with bigger timescales through a carefully designed decaying strategy. By doing so, the model explores the differences between the training and inference phases as well as the intermediate model in order to correct its exposure bias. Experimental results demonstrate our model achieves superior performance as well as faster convergence time on two spatio-temporal sequence forecasting datasets. Experiments also show that our proposed method is better at modelling long term dependencies hence our method produces better long term prediction accuracies.

Main contributions of this paper are summarised as follows:
\begin{itemize}
\item We propose a novel temporal progressive growing sampling method to effectively bridging the gap between training and inference for spatio-temporal forecasting. Model is trained with bigger time gap initially and gradually transform into smaller time gap.

\item We carefully design a decay strategy that take account of current index of the sequence, the latter sequence with larger probability to be replaced by the generated predictions during training, which helps the convergence of the training and yield better performance.

\item We conduct extensive experiments on two real-world spatio-temporal datasets to evaluate the performance of our proposed method. Experimental results reveal that our approach achieves superior performance on sequence forecasting tasks, and experimental results also show our approach achieve better long term prediction accuracies.
\end{itemize}

The rest of paper is organised as follows. Section~\ref{sec:model} introduces our proposed \modelshort and discusses the details of our model. Section~\ref{sec:exp-wf} and  Section~\ref{sec:exp-mm} present the experimental results of weather prediction, and with Moving MNIST dataset~\cite{Srivastava:2015:ULV:3045118.3045209}. Then Section~\ref{sec:related} draws links to some related studies. Lastly, Section~\ref{sec:conclusion} presents concluding remarks and lists several directions for future work.

\section{\model} \label{sec:model}

\subsection{Seq2Seq and Scheduled Sampling}
Sequence to Sequence model (Seq2Seq) \cite{NIPS2014_5346} was first introduced to solve complex sequential problems that traditional RNN approaches cannot model diverse input and output lengths. Seq2Seq is also known as an encoder-decoder where the encoder encodes the original sequence into a feature vector, then the decoder outputs a target sequence based on the feature vector and previous contexts. Typically, both encoder and decoder are RNN, and Seq2Seq has been used for sequence modeling tasks like NMT, speech recognition and recently for STSF tasks \cite{li2018diffusion,2017arXiv170404110S,2017arXiv171100073Y}.

One of the main drawbacks of Seq2Seq models is that, small errors can accumulate quickly along the generated sequence at the inference stage due to the different distributions of training and inference phases. That is because RNN models minimise single step errors only during training when all previous ground-truth contexts are available. However, the entire sequence has to be generated during the inference phase which causes a discrepancy between training and inference. Bengio \textit{et al.} proposed a sampling strategy called Scheduled Sampling to close the gap between training and inference for NMT, which is explained as below:

\begin{equation}
\label{ss}
\begin{array}{l}
{\forall~k, 1 \leq k \leq K,} \\ 

{\hat{\mathcal{X}}_{t+k} \sim \mathcal{M}\left(\mathrm{Encoder}(\mathcal{S}), \tilde{\mathcal{X}}_{t+1 : t+k} ; \theta\right),} \\ 

 {\tau_{t+k+1} \sim \mathrm{Bernoulli}\left(1, \epsilon_{i}\right)},\\

{\tilde{\mathcal{X}}_{t+k+1}=\left(1-\tau_{t+k+1}\right) \hat{\mathcal{X}}_{t+k}+\tau_{t+k+1} \mathcal{X}_{t+k}.} 

 \end{array}
\end{equation}

\noindent Here, $\mathcal{M}\left(\mathrm{Encoder}(\mathcal{S}), \tilde{\mathcal{X}}_{t+1 : t+k} ; \theta\right)$ denotes one step prediction based on the encoder and the previous inputs $\tilde{\mathcal{X}}_{t+1 : t+k}$. $\tau_{t+k+1} $ is a random variable generated by a coin flip following the Bernoulli distribution where $\epsilon_{i}$ is the probability of $\tilde{\mathcal{X}}_{t+k+1}$ sampling from the ground-truth $\mathcal{X}_{t+k}$, i.e., $1-\epsilon_{i}$ is the probability of sampling from the previous prediction $ \hat{\mathcal{X}}_{t+k}$. When $\epsilon_{i} = 1$, $\tilde{\mathcal{X}}_{t+k+1}$ is always sampling from the ground-truth $\mathcal{X}_{t+k}$. When $\epsilon_{i} = 0$, $\tilde{\mathcal{X}}_{t+k+1}$ is always sampling from the previous output $ \hat{\mathcal{X}}_{t+k}$. During training, the probability $\epsilon_{i}$ is decreased from 1 to 0.

\subsection{Bridging the Gap with TPG Sampling}

%The basic idea of \emph{curriculum learning} is to start learning easier tasks first, then gradually increases learning difficulties \cite{ELMAN199371,Bengio:2009:CL:1553374.1553380}. Progressive growing training strategy was introduced in \cite{karras2018progressive} to overcome the training difficulties in GAN models. The idea is start training the generator and discriminator with small resolutions of images and gradually transforms them to bigger resolutions. The underlying principle of progressive growing is the same as \emph{curriculum learning}, which is to \emph{start small}. While Schedule Sampling overcomes the training difficulties of long term dependencies of Seq2Seq model, we want to take one step further to take advantage of the progressive growing idea along with Schedule Sampling to overcome facing problems in STSF.

While adopting Seq2Seq from NMT to STSF brings significant benefits, it also creates some drawbacks. The discrepancy between training and inference also creates a gap in STSF systems. However, unlike NMT systems that the gap might result in a complete different translation, errors at previous steps in STSF could generate confusion to the local trends. Therefore, an unchanged adoption of Scheduled Sampling to STSF is not an ideal solution. Furthermore, spatio-temporal dynamics can be modeled by different sampling rates. For instance, if we assume that we sample data every 1 hour, then we can train a model to estimate the prediction of every 2 hours by feeding the odd index sequence and the even index sequence separately.  Current Scheduled Sampling method does not consider such unique characteristic of STSF.

 \begin{figure}[!h]
	\centering
	\includegraphics[width=0.42\textwidth, height=0.76\textwidth]{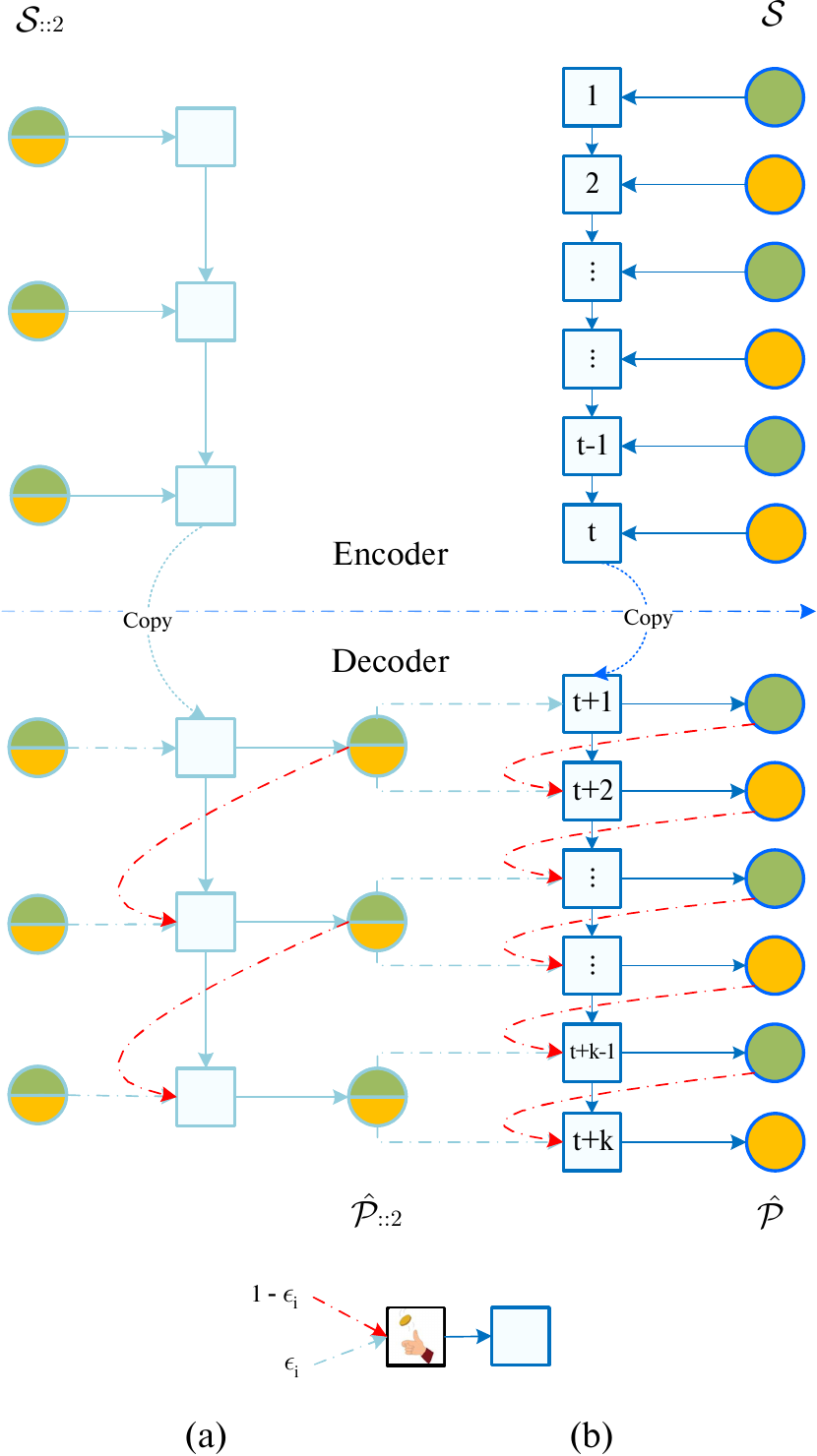}
	\caption{Illustration of our proposed \modelshort sampling. Green circles represent odd index inputs $\mathcal{X}_\mathrm{odd}$ and outputs $\hat{\mathcal{X}}_\mathrm{odd}$, orange circles represent even index inputs $\mathcal{X}_\mathrm{even}$ and outputs $\hat{\mathcal{X}}_\mathrm{even}$. When start training $\mathcal{M}_1$ (a)  model initially, each sequence is fed with a subsequence of green and orange. During the transition to model $\mathcal{M}_2$ (b), the decoder input $\tilde{\mathcal{X}}_{t+\frac{k}{2}::2}$ of model $\mathcal{M}_1$ is used as a source for sampling.}
	\label{fig:tp}
\end{figure}

Motivated by this, we propose a \modelshort sampling strategy to close the gap between training and inference. First, we subsample the sequence into two subsequences by separating the odd and even indexes (see Fig. \ref{fig:tp}, green colour represents the odd index inputs and orange represents the even). The idea is to start training a simple model $\mathcal{M}_1(\mathrm{Encoder}(\mathcal{S}_{::2}), \theta)$ that takes the odd or even index sequence denoted as $\mathcal{S}_{::2}$. The benefit of this approach is that the sequence length is cut to half to the original length, which is much easier for both the encoder and decoder to learn.  Moreover, model $\mathcal{M}_1$ is trained with Scheduled Sampling as in Equation \ref{ss}, then the decoder input of model $\mathcal{M}_1$ is used as a sampling source for model $\mathcal{M}_2$ during the transition phase. This brings another advantage that the model is not only able to explore the differences between training and inference from itself but also the intermediate model trained with larger time scales. The detailed transition process is described as follows:

 \begin{equation}
\begin{array}{l}
{\forall~k, 1 \leq k \leq K,} \\ 

{\hat{\mathcal{X}}_{t+\frac{k}{2}::2} \sim \mathcal{M}_1\left(\mathrm{Encoder}_1(\mathcal{S}_{::2}), \tilde{\mathcal{X}}_{t+1 : t+\frac{k}{2}::2} ; \theta_1\right),} \\ 

{\hat{\mathcal{X}}_{t+k} \sim \mathcal{M}_2\left(\mathrm{Encoder}_2(\mathcal{S}), \tilde{\mathcal{X}}_{t+1 : t+k} ; \theta_2\right),} \\ 

 {\tau_{t+k+1} \sim \mathrm{Bernoulli}\left(1, \epsilon_{i}\right)},\\

{\tilde{\mathcal{X}}_{t+k+1}=\left(1-\tau_{t+k+1}\right) \hat{\mathcal{X}}_{t+k}+\tau_{t+k+1} \tilde{\mathcal{X}}_{t+\frac{k}{2}::2}. }

 \end{array}
\end{equation}
 \noindent Here, $\mathcal{X}_{{t}+\frac{k}{2}::2}$ denotes decoder inputs from model $\mathcal{M}_1$ of the corresponding index of model $\mathcal{M}_2$. Similar to Scheduled Sampling, $\epsilon_{i}$ is the probability that follows the Bernoulli distribution that controls whether $\tilde{\mathcal{X}}_{t+k+1}$ samples are from the previous output $\hat{\mathcal{X}}_{t+k}$ or $\tilde{\mathcal{X}}_{{t}+\frac{k}{2}::2}$ of $\mathcal{M}_1$. When $\epsilon_{i} = 1$, $\tilde{\mathcal{X}}_{t+k+1}=\hat{\mathcal{X}}_{t+k}$. During the transition from $\mathcal{M}_1$ to $\mathcal{M}_2$, we decrease $\epsilon_{i}$ from 1 to 0. When $\epsilon_{i} = 0$, then model $\mathcal{M}_2$ gets solely trained conditioned on its previous output. 
 
 Although our proposed TPG Sampling share some similarities to Scheduled Sampling, there are two key differences. First, TPG closes the gap between training and inference not only from exploring the bias but also corrects the errors caused by the bias by learning the intermediate model of large time scale. Second, we careful design a decaying strategy to work with TPG which is introduced in the next section, our experiments show significant improvements. 
  
 \subsection{Decay Strategy}
 
 Scheduled Sampling decreases $\epsilon_{i} $ during the transition period by the inverse sigmoid function as follows:

  \begin{equation}
  	\epsilon_{i} = \frac{\lambda}{\lambda + \exp(i / \lambda)}
  \end{equation}
where $i$ is the current global batch number, $\lambda$ is the parameter setting to control the deceasing speed of $\epsilon_{i}$ and therefore the convergence speed. Here, the whole sequence shares the same probability to replace ground-truth with output generated by model itself, noted $\epsilon_{i}$ towards 0 the greater probability is. However, during our experiment we observed that STSF model converges from the begin of the sequence. Replacing ground-truth with system generated output for the begin of the sequence increases the convergence difficulty of the whole training process. Therefore, we propose a decay strategy that takes the current index of the sequence into account when calculating the probability as described as follows:
% return k / (k + np.exp(global_step * np.log(i+2) / k))
  \begin{equation}
  	\epsilon_{i}^{v} = \frac{\lambda}{\lambda + \exp(i \times \log(\upsilon) / \lambda)}
  \end{equation}
where $\upsilon$ is the current index of the sequence starting from 2. Therefore, later index input has a bigger probability where ground-truth values are replaced than the earlier input at the beginning of the training process. This helps the convergence of training by keeping previous input as ground truth. However, $\epsilon_{i}^{v}$ from different indexes will become smaller and eventually to zero due to the exponential grow as towards to the end of the training, which is desire because we want all ground truth to be replaced by model generated outputs when the training is finished.

\section{Weather Forecasting Experiments} \label{sec:exp-wf}

To evaluate advances of our TPG sampling, we first conducted experiments on a weather forecasting task. Weather forecasting is a typical STSF task that brings many benefits to people's everyday life as well as agriculture and many more. Experimental results show our method outperforms several baseline methods as well as the basic Seq2Seq with Scheduled Sampling.  
\subsection{Dataset}

\begin{figure*}[h]
%\vspace{0.1cm}
  \begin{minipage}[b]{0.65\textwidth}
  \tiny
    \centering
    \resizebox{1\textwidth}{!}{%
    \begin{tabular}{lcccccc}
    \toprule[0.15ex]
        & \multicolumn{2}{c}{\textbf{t2m}} & \multicolumn{2}{c}{\textbf{rh2m}} & \multicolumn{2}{c}{\textbf{w10m}} \\ 
    \toprule[0.15ex]
        & RMSE        & MAE    & RMSE        & MAE     & RMSE        & MAE    \\ 
	\midrule[0.1ex]
	NWP     &     2.939        &    2.249  &         18.322        &  13.211    &           1.813         &  1.327         \\ 
	ARIMA   &   3.453          &  2.564    &      18.887             &   14.163    &   2.436         &   1.675             \\ 
	Seq2Seq $_{\scaleto{180}{2pt}}$ without sampling  &       3.149      &  2.393          &      16.230           &   11.712         &      1.437        &  1.032       \\ 
	Seq2Seq $_{\scaleto{90}{2pt}}$  + Scheduled Sampling &        2.828       &     2.173       &       16.023        &    11.428     &       1.380       &   0.962        \\ 
	Seq2Seq $_{\scaleto{180}{2pt}}$ + Scheduled Sampling   &       3.080      &  2.334          &       \textbf{14.180}           &   \textbf{10.254}         &      1.417        &  1.035       \\ 
	\modelshort$_{\scaleto{\mathcal{M}_1}{2.5pt}}$     &    3.006   &      2.379             &     16.639    &         12.197                &   2.211   &   1.360         \\ 
	\modelshort     &       \textbf{2.611}       &    \textbf{1.984}        &      14.994         &   10.623         &    \textbf{1.328}            &      \textbf{0.914}      \\ 
	\bottomrule[0.15ex]
	\end{tabular}}

      \captionof{table}{Experimental results based on the test set from 01-06-2018 to 28-08-2018. Smaller number indicates smaller prediction errors.}
       \label{Tab:results-wf}
    \end{minipage}
    \begin{minipage}[b]{0.35\textwidth}
    \centering
    \includegraphics[width=0.93\textwidth]{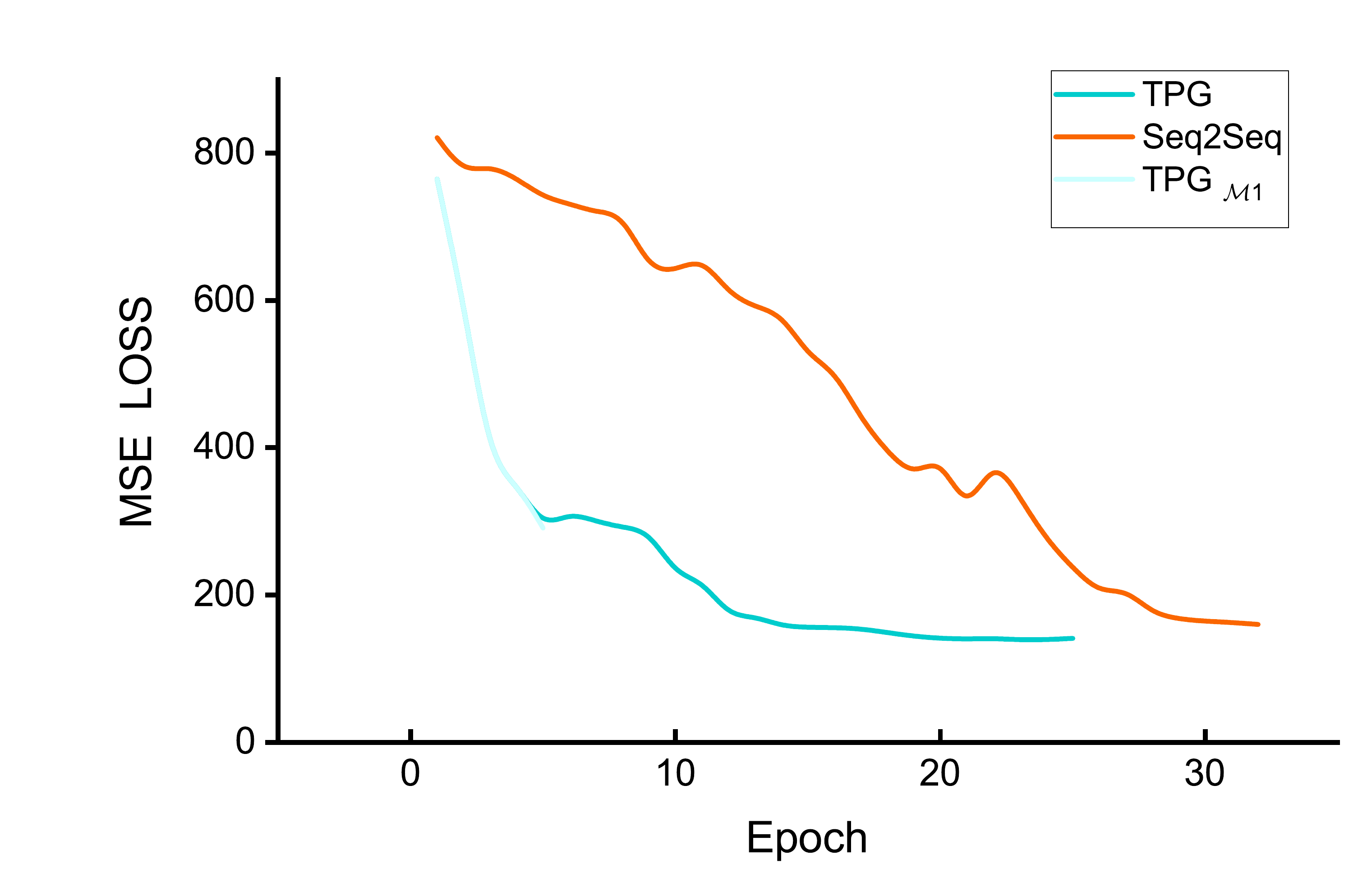}
    \captionof{figure}{Test set loss comparison during training.}
        \label{Figure:wf-loss}
  \end{minipage}
%\vspace{0.1cm}
  \end{figure*}

AI Challenger weather forecasting is an online competition \footnote{https://challenger.ai/competition/wf2018}, and the goal is to predict air temperature at 2 meter (t2m), relative humidity at 2 meter (rh2m) and wind speed at 10 meter (w10m) across 10 weather stations in Beijing city. This dataset contains historic observations from 01-03-2015 to  30-10-2018. We use 01-03-2015 to 31-05-2018 for our training set, 01-06-2018 to 28-08-2018 for the test set, and 29-08-2018 to 30-10-2018 for validation purpose. Each node contains 9 measurements including t2m, rh2m, w10m and 6 others, as well as 37 other predictions which are generated by the Numerical Weather Prediction (NWP). 

%\begin{figure}[!thbp]
%	\begin{subfigure}[b]{0.5\textwidth}
%	\centering
%		\includegraphics[width=\textwidth]{assets/t2m-obs.png}
%		\label{sfig:t2m}
%		\vspace{-0.4cm}
%		\caption{t2m}
%	\end{subfigure}\hfill
%	\begin{subfigure}[b]{0.5\textwidth}
%	\centering
%	   \vspace{0.3cm}
%		\includegraphics[width=\textwidth]{assets/rh2m-obs.png}
%		\label{sfig:rh2m}
%		\vspace{-0.4cm}
%		\caption{rh2m}
%	\end{subfigure}\hfill
%	\begin{subfigure}[b]{0.5\textwidth}
%	\centering
%	\vspace{0.3cm}
%		\includegraphics[width=\textwidth]{assets/w10m-obs.png}
%		\label{sfig:w10m}
%		\vspace{-0.4cm}
%		\caption{w10m}
%	\end{subfigure}\hfill
%	\vspace{0.5cm}
%	\caption{Example data plots from 01-03-2015 to 01-06-2016 of t2m, r2hm and w10m, where t2m shows very strong season trends, r2hm shows medium season trends and w10m has no season trends.}
%	\label{Fig:plot}
%\end{figure}
%\vspace{0.5cm}

%We can see from Fig. \ref{Fig:plot} that air temperature (t2m) has very strong season trends and daily recurrent patterns, humidity (r2hm) has medium season trends and daily recurrent patterns, whereas wind speed has almost no obvious pattern which makes it harder to predict. Our task is to learn such trends and patterns based on the historic observations and to make predictions for future trends.

\subsection{Implementation}

%\begin{table*}[!ht]
%\centering
%\begin{tabular}{ccccccc}
%\hline
%        & \multicolumn{2}{c}{t2m} & \multicolumn{2}{c}{rh2m} & \multicolumn{2}{c}{w10m} \\ \hline
%        & RMSE        & MAE    & RMSE        & MAE     & RMSE        & MAE    \\ \hline
%NWP     &     2.939        &    2.249  &         18.322        &  13.211    &           1.813         &  1.327         \\ 
%ARIMA   &             &      &                   &       &            &                \\ 
%Seq2Seq $_{90}$ &        2.828       &     2.173       &       16.023        &    11.428     &       1.380       &   0.962        \\ 
%Seq2Seq $_{180}$   &       3.080      &  2.334          &       14.180           &   10.254         &      1.417        &  1.035       \\ 
%\modelshort$_{m1}$     &    3.006   &      2.379             &     16.639    &         12.197                &   2.21   &   1.360         \\ 
%\modelshort$_{\lambda1=1000, \lambda2=1000}$     &      2.890        &     2.265       &        15.174          &    10.851        &       1.448       &    1.058      \\ 
%\modelshort$_{\lambda1=500, \lambda2=1000}$     &       2.611       &    1.984        &      14.994         &   10.623         &    1.328            &      0.914      \\ \hline
%\end{tabular}
%\caption{Experimental results based on test set from 01-06-2018 to 28-08-2018}
%\label{tab:wf-results}
%\end{table*}

\subsubsection{Seq2Seq}
As mentioned previously, we have 10 weather stations, each station contains observations of 9 different measurements. It is the challenge to model correlations between stations as well as correlations between different measurements. For example, wind speed and air temperature have strong correlations. In fact. every measurement could have impact on each other, however it is non-obvious but highly dynamic. As stated in the previous section, our approach is based on Seq2Seq which is capable of  modeling such correlations with its encoder and decoder architecture. First, we use a LSTM encoder taking an input $\mathcal{S} \in \mathbb{R}^{T\times10\times9} $ which produces a feature vector $\mathbf{h} \in \mathbb{R}^C$. Here, $T$ is a hyper-parameter to specify the sequence length we use for feature encoding, 10 is the total number of stations, and 9 means we use all 9 measurements for feature learning. This process is called feature extraction or feature learning. During the computation of LSTM encoder for each time step, the linear transformation operation takes account of 9 measurements of 10 stations. The overall feature of all time steps will then be encoded into one vector space $\mathbf{h}$. Then such feature vector $\mathbf{h}$ will be decoded by a LSTM decoder to produce an output $\hat{\mathcal{S}} \in \mathbb{R}^{37\times 10 \times 9}$ which is the following 37 hours prediction of 9 measurements. Note that, we are only predicting t2m, rh2m and w10m. Therefore, we follow a fully-connected layer to output the final prediction $\hat{\mathcal{P}} \in \mathbb{R}^{37\times 10 \times 3} = \hat{\mathcal{S}} \mathbf{W}_s+\mathbf{b}_s$.

\subsubsection{\modelshort}
We extend the base Seq2Seq model with our proposed \modelshort sampling. To be more specific, the source sequence of length 37 is separated half into odd index and even index sequences of length 19 and 18, respectively. Such sequences are then used for the initial training of model $\mathcal{M}_1$. We turn the network with different settings of $\lambda_1$ and $\lambda_2$ and report comparative results to see the impact of different settings in Section~\ref{sec:evaluation}.
\subsubsection{Loss Function}

We divide the loss function into three parts:
\begin{equation}
	\mathcal{L} = \mathrm{MSE}_{t2m} + \mathrm{MSE}_{rh2m} + \mathrm{MSE}_{w10m},
\end{equation}
where MSE denotes \emph{mean squared error}: $\mathrm{MSE} = \frac{1}{n} \sum_{i=1}^{n}\left(\mathbf{X}_{i}-\hat{\mathbf{X}}_{i}\right)^{2}$. During training, $\mathcal{L}$ is optimised jointly by BPTT \cite{58337} with Adam optimiser \cite{2014arXiv1412.6980K}.

\subsubsection{Parameter Settings}
For encoder and decoder, we choose a dimension $C = 90$ for LSTM whilst for encoder, $T$ is set to 96 which means we use 4 days make up of 96 hours historic observations to predict the next 37 hours. For \modelshort, $\lambda = 3000$ for weather forecasting and $\lambda = 1000$ for Moving MNIST++. Learning rate for Adam is set to $1e^{-2}$. 

\subsubsection{Experiment Environment}
We implement our model using Tensorflow 1.12, a well known deep learning library developed by Google. Our model is trained and evaluated on a server with Nvidia V100 GPU and Intel(R) Xeon(R) Gold 5118 CPU @ 2.30GHz (24 cores).

\subsection{Overall Evaluation}\label{sec:evaluation}

\subsubsection{Evaluation Matrix}

We report experimental results on each measurement for the test set using \emph{Root Mean Squared Error} (RMSE) and \emph{Mean Absolute Error} (MAE):
% and \emph{Mean absolute percentage error} (MAPE):

\begin{equation}
	\mathrm{RMSE} = \sqrt{ \frac{1}{n} \sum_{i=1}^{n}\left(\mathbf{X}_{i}-\hat{\mathbf{X}}_{i}\right)^{2}}.
\end{equation}

\begin{equation}\label{equation:mae}
	\mathrm{MAE} =\frac{1}{n}{\sum_{i=1}^{n}\left|\mathbf{X}_{i}-\hat{\mathbf{X}}_{i}\right|}.
\end{equation}

Here, $n$ is the number of prediction time steps times the number of locations.

%\begin{figure}[!hbp]
%	\centering
%	\includegraphics[width=0.45\textwidth]{assets/wf-training}
%	\caption{Test set loss comparison during training.}
%	\label{fig:tp}
%\end{figure}
\begin{figure*}[h]
\centering
   \begin{subfigure}{0.48\linewidth}
   \centering
  \includegraphics[width=3.6in]{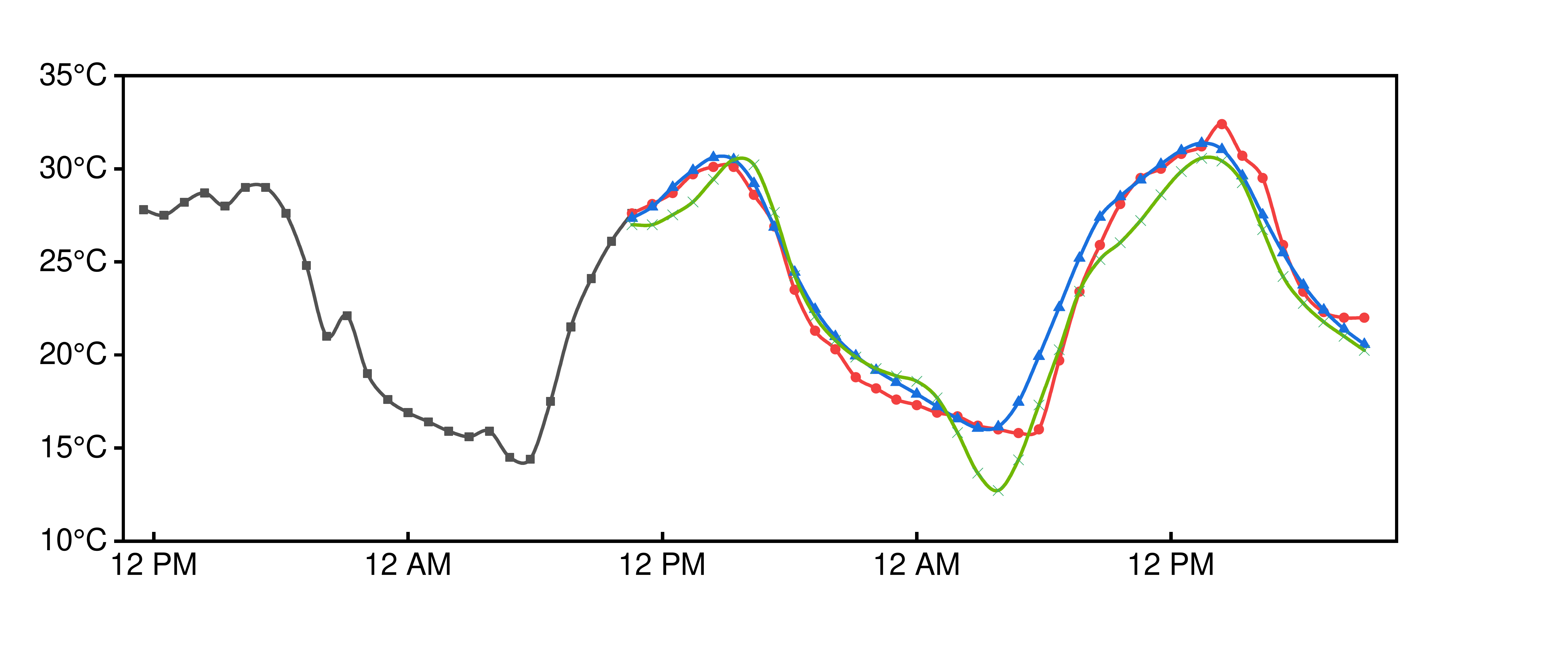} 
   \caption{t2m predictions}
   \label{sfig:t2m-example} 
\end{subfigure}
\hfill
\begin{subfigure}{0.48\linewidth}
   \centering
   \includegraphics[width=3.6in]{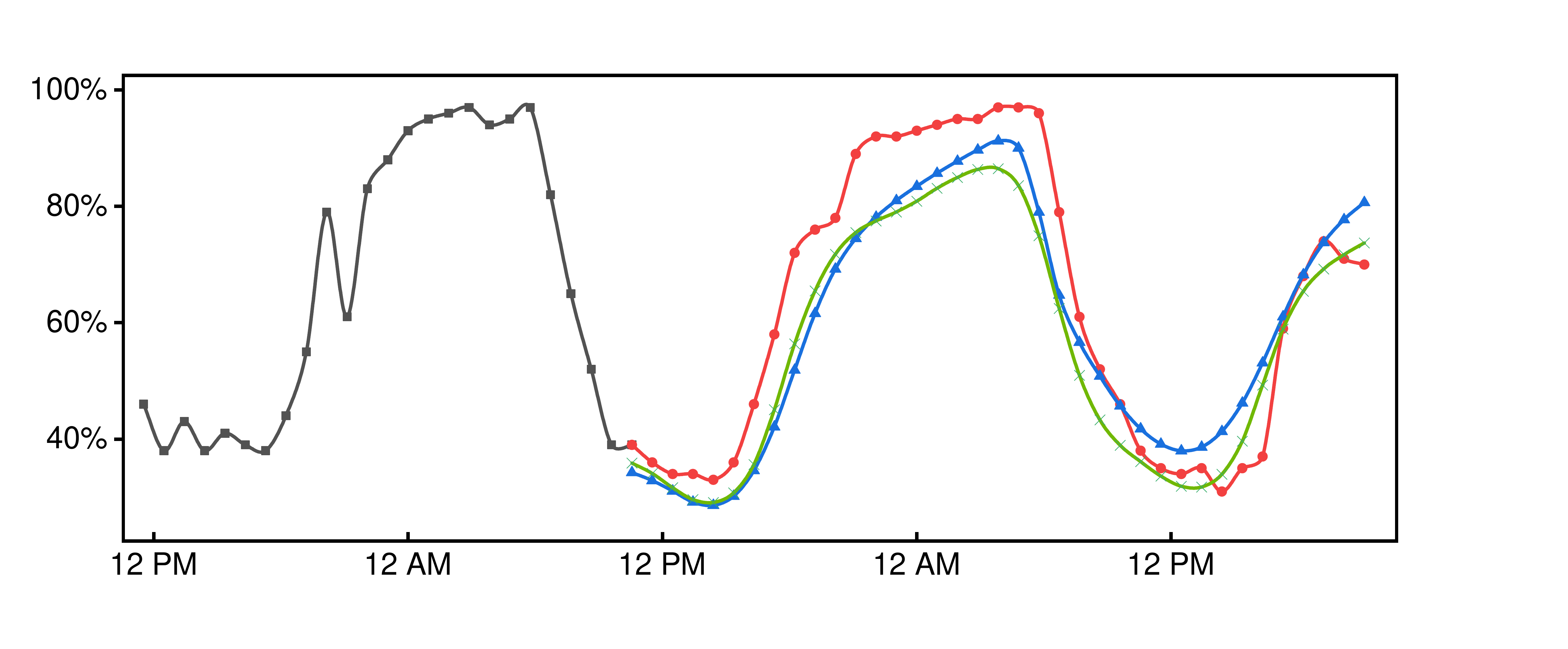} 
   \caption{rh2m predictions}
   \label{sfig:rh2m-example}
\end{subfigure}
\\[\baselineskip]
\begin{subfigure}[H]{0.58\linewidth}
   \centering
   \includegraphics[width=4.5in]{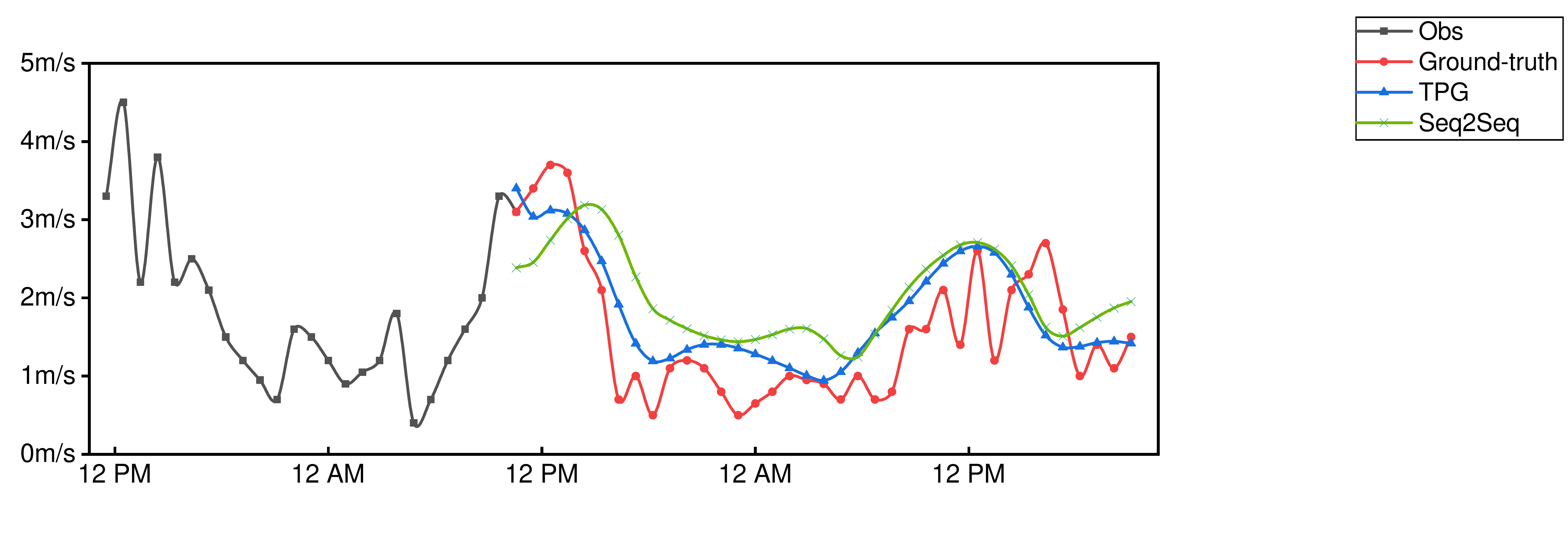} 
   \caption{w10m predictions}
   \label{sfig:w10m-example}
\end{subfigure}
\centering
\caption{Prediction visualisations (based on 30-08-2018) of air temperature (t2m), relative humidity (rh2m) and wind speed (w10m). \modelshort predictions have less outliers and have more accurate long term predictions.}
\label{fig:pred-example}
\end{figure*}

\begin{figure*}[h]
   \begin{subfigure}{0.4\linewidth}
   \vspace{4.8cm}
   \includegraphics[width=0.8\linewidth]{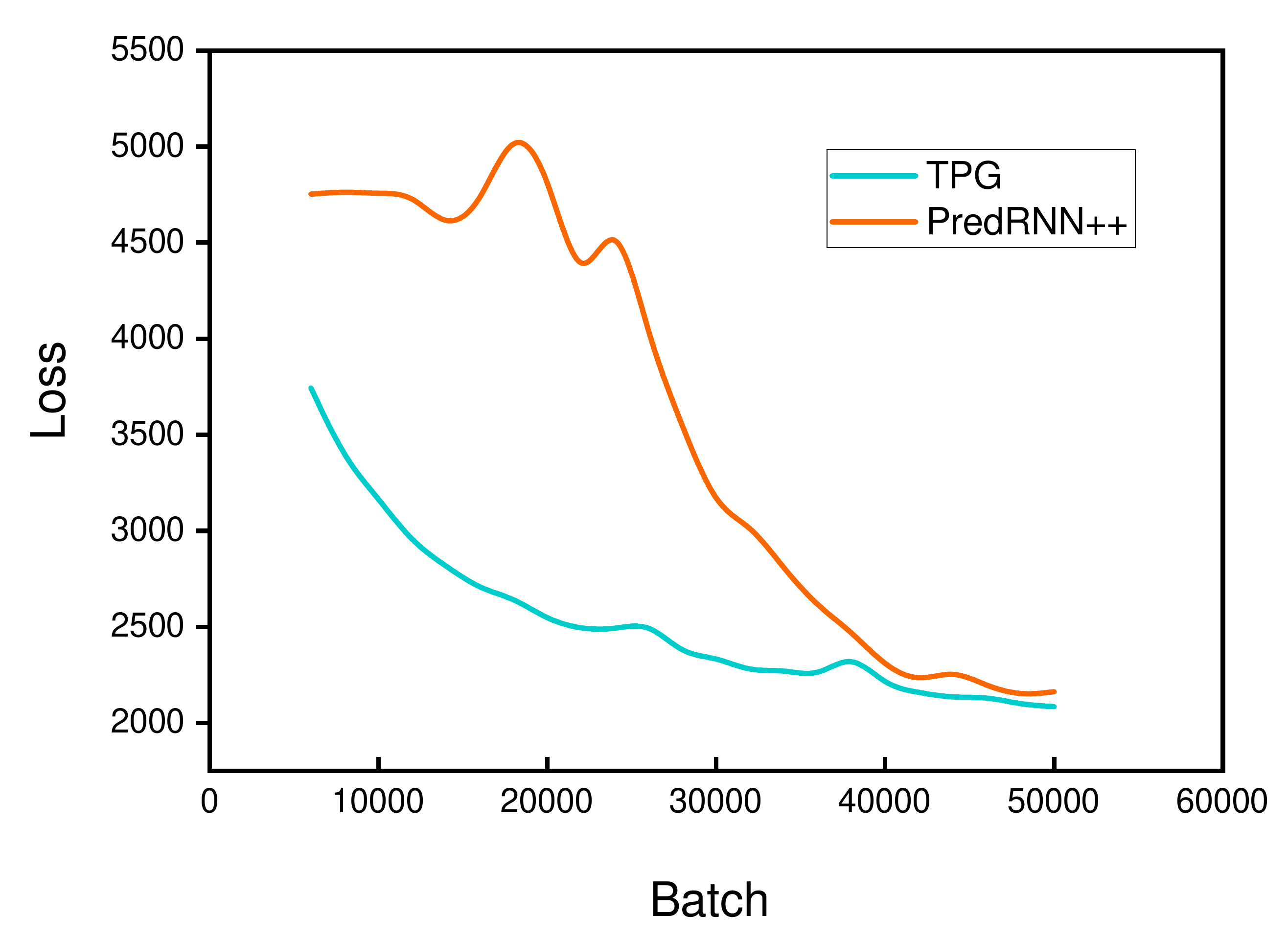} 
   \centering
    \vspace{0.3cm}
   \caption{Test set loss during training}
   \label{sfig:mm-training} 
\end{subfigure}
\begin{subfigure}{0.6\linewidth}
   \hspace{-4.5cm}\includegraphics[width=1.3\linewidth]{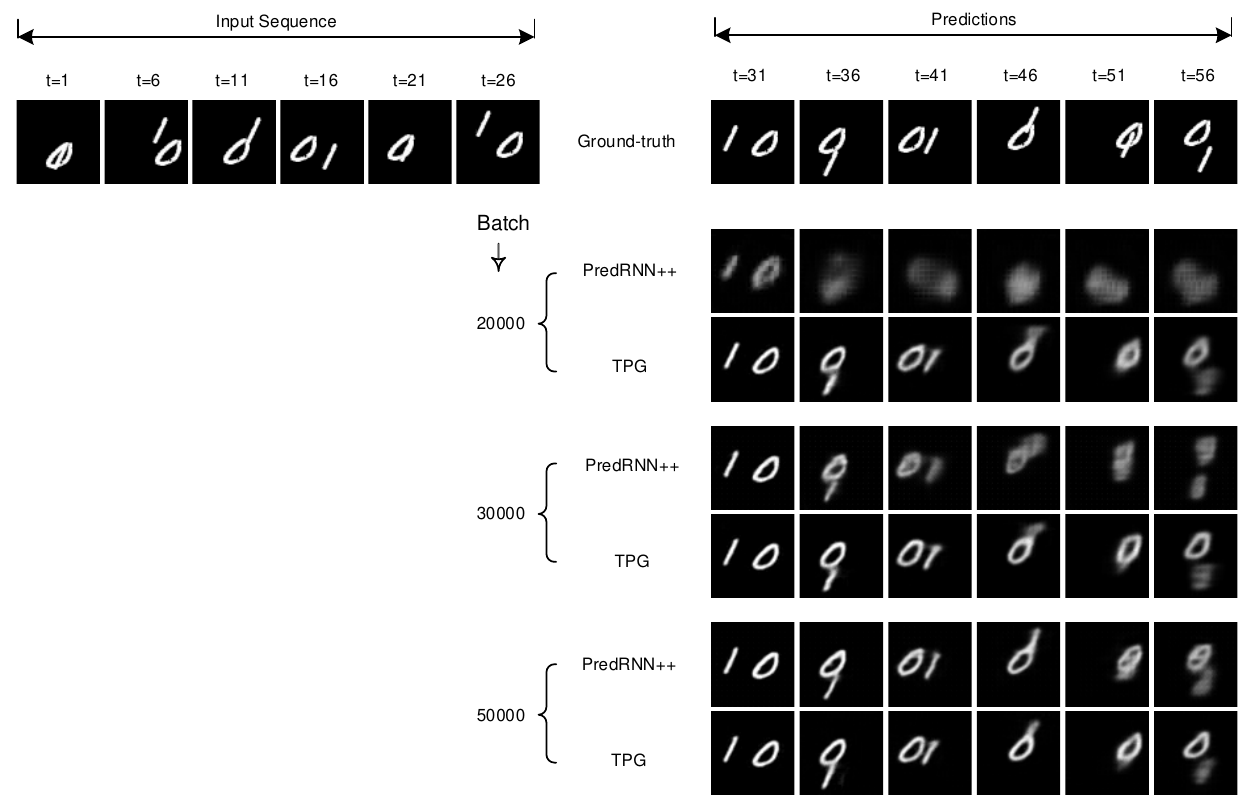} 
   \vspace{0.4cm}
\caption{Moving MNIST prediction visualisations during training.}
   \label{sfig:mm-example}
   
\end{subfigure}

\centering
  \vspace{0.4cm}
\caption{A visualisation of training progress: (\ref{sfig:mm-training}) shows our proposed \modelshort converges faster during training; (\ref{sfig:mm-example}) shows our proposed \modelshort trains long range predictions faster and have more accurate long range predictions.}
\label{fig:mm}
\end{figure*}

\hfill
\subsubsection{Baselines}
\hfill
\par\smallskip
\noindent \textbf{NWP}:  Numerical Weather Prediction~\cite{Lynch:2008:OCW:1347465.1347773} uses mathematical models of the atmosphere and oceans to predict the weather based on current weather conditions. 
\par\smallskip

\noindent \textbf{ARIMA}:  Auto-Regressive Integrated Moving Average is the most common baseline method for time series predictions \cite{Box:1990:TSA:574978}. Model is trained individually by each station.
\par\smallskip

\noindent \textbf{Seq2Seq $_{180}$} without sampling:  Basic Seq2Seq model with the LSTM dimension is set to 180. Other parameters are set the same as \modelshort model.
\par\smallskip

\noindent \textbf{Seq2Seq $_{90}$ } + Scheduled Sampling:  Basic Seq2Seq model with Scheduled Sampling of the LSTM dimension is set to 90. Other parameters are set the same as \modelshort model.
\par\smallskip

\noindent \textbf{Seq2Seq $_{180}$} + Scheduled Sampling:  Basic Seq2Seq model with Scheduled Sampling of the LSTM dimension is set to 180. Other parameters are set the same as \modelshort model.
\par\smallskip

\noindent \textbf{\modelshort$_{\mathcal{M}_1}$}: Proposed \modelshort model with $\lambda$ is set to 500, results reported based on the intermediate model $\mathcal{M}_1$.
\par\smallskip

\noindent \textbf{\modelshort}: Proposed \modelshort model with $\lambda$ is set to 500, results reported based on the final model $\mathcal{M}_2$.
\par\smallskip

Note that all deep learning based models are trained with the same parameter settings otherwise stated above. Models are trained with training set,  the best models are chosen which demonstrate the best performance based on the validation set, and experimental results are reported based on the test set.

\subsubsection{Performance Evaluation}

We compare our \modelshort to the baselines listed above including the tradition mathematical model NWP, and machine learning based approaches. Table \ref{Tab:results-wf} shows a summary of experimental results based on two evaluation matrices. It clearly demonstrates that our \modelshort outperforms all baselines for three measurements under study.  

First, Seq2Seq based models outperform traditional approaches including NWP and ARIMA. Seq2Seq based models utilise the LSTM encoders and decoder which effectively model the spatio-temporal dynamics. To be more specific, Seq2Seq $_{90}$ performs better than Seq2Seq $_{180}$ in overall, but in particular for t2m and w10m. 

Second, Table \ref{Tab:results-wf} also shows the Seq2Seq model trained with Scheduled Sampling performs better than Seq2Seq without sampling which shows Scheduled Sampling improves STSF. Our proposed TPG sampling continue to improve the effectiveness of bridging the gap between training and inference.

Third, our proposed \modelshort model outperforms base Seq2Seq models. The test of significance in terms of both accuracy measures shows that \modelshort significantly improves Seq2Seq $_{90}$. As shown in Fig. \ref{Figure:wf-loss}, our proposed \modelshort model converges significantly faster than basic Seq2Seq models. The red line represents the test loss of Seq2Seq model, whilst the light blue line represents the test loss of model $\mathcal{M}_1$, and the dark blue line represents the test loss of model $\mathcal{M}_2$. Benefits from shorter sequence length, model $\mathcal{M}_1$ quickly converges compared to original Seq2Seq. Resulting in the second stage of training model $\mathcal{M}_2$, $\mathcal{M}_2$ shows faster convergence as well as better final performance. 

%Last, Fig. \ref{fig:pred-example} shows visualisations of predictions of Seq2Seq and \modelshort models for a particular day. As shown in Fig. \ref{sfig:t2m-example}, \modelshort model has less outlier predictions in t2m predictions. Visualisations also show that our model is better at modeling long term dependencies where they show better long term prediction accuracies for the three measurements.

To sum up, our proposed \modelshort archives the best overall performance. To be specific, \modelshort converges much faster during training due to the progressive growing training mechanism. Prediction examples also show that \modelshort predictions are more reliable on the outlier predictions and the long range predictions than baseline approaches and Scheduled Sampling.

\section{Moving MNIST Experiments} \label{sec:exp-mm}
The weather forecasting dataset shows the outperforming performance of our proposed \modelshort model on a vector like dataset. In order to further evaluate the usability and applicability of our proposed method on image-like spatio-temporal sequential datasets, we further conduct experiments on the Moving MNIST dataset \cite{Srivastava:2015:ULV:3045118.3045209}.
\subsection{Dataset}
The Moving MNIST dataset is originally for evaluating video (a series of sequential images) prediction performance, since then it becomes one of the most common spatio-temporal sequence prediction benchmarks \cite{Shi:2015:CLN:2969239.2969329,NIPS2017_6689,pmlr-v80-wang18b}. We generate a series of image sequences containing two moving handwritten digits with different moving speed and velocity. In addition to the typical setup, we extend the sequence length to 60: 30 for the input sequence and another 30 for prediction. We generate 10,000 sequences for training, 3,000 for validation and 5,000 for testing. Experimental results are reported based on the test dataset.
\subsection{Implementation}

\subsubsection{Seq2Seq} We use open source code\footnote{https://github.com/Yunbo426/predrnn-pp} provided by PredRNN++ \cite{pmlr-v80-wang18b} for our base Seq2Seq model for images. PredRNN++ is a strong baseline for video frame prediction which is based on Seq2Seq with a custom RNN cell, called Casual LSTM and a GHU unit. 

\subsubsection{\modelshort} Then we extend the baseline Seq2Seq model with our proposed \modelshort. We choose $\lambda = 3000$ and we train the model for 50000 iterations.

\subsection{Overall Evaluation}

\subsubsection{Evaluation Matrix}
We report results based on two matrices: per-frame Structural Similarity Index Measure (SSIM) \cite{1284395} and MSE. Larger SSIM scores indicate greater similarities between the ground-truth and prediction whilst smaller MSE values indicate smaller prediction errors.

\subsubsection{Baselines}
We compare the performance of our approach against several baseline methods as in PredRNN++ \cite{pmlr-v80-wang18b} including ConvLSTM \cite{Shi:2015:CLN:2969239.2969329}, TrajGRU \cite{xingjian2017deep}, CDNA \cite{Finn:2016:ULP:3157096.3157104}, DFN \cite{DeBrabandere:2016:DFN:3157096.3157171}, VPN \cite{Kalchbrenner:2017:VPN:3305381.3305564} and PredRNN \cite{NIPS2017_6689}.

\begin{table}
\centering
\begin{tabularx}{0.45\textwidth}{p{5cm}p{1cm}p{1cm}}
\toprule[0.15ex]
     & SSIM  $\uparrow$  & MSE $\downarrow$    \\ 
\toprule[0.15ex]
%FC-LSTM \cite{Srivastava:2015:ULV:3045118.3045209}     &      0.583       &   180.1     \\ 
ConvLSTM \cite{Shi:2015:CLN:2969239.2969329}   &    0.597         &     156.2      \\ 
TrajGRU \cite{xingjian2017deep} &     0.588          &     163.0         \\ 
CDNA \cite{Finn:2016:ULP:3157096.3157104}   &    0.609         &    142.3          \\ 
DFN \cite{DeBrabandere:2016:DFN:3157096.3157171}     &  0.601     &      149.5                 \\ 
VPN \cite{Kalchbrenner:2017:VPN:3305381.3305564}     &     0.620         &      129.6      \\ 
PredRNN \cite{NIPS2017_6689}    &    0.645          &   112.2         \\
\midrule[0.07ex]
PredRNN++ without sampling &    0.733          &   91.10         \\
PredRNN++ with SS \cite{pmlr-v80-wang18b}    &    0.769          &   87.74        \\
\modelshort &    \textbf{0.811}          &   \textbf{85.41}   \\
\bottomrule[0.15ex]
\end{tabularx}
\caption{Experimental results of 30 time steps prediction: results reported for per frame. Baselines reported as in PredRNN++ \cite{pmlr-v80-wang18b}.}
\label{tab:mm-results}
\end{table}

\subsubsection{Performance Evaluation}
Table \ref{tab:mm-results} shows a summary of experimental results with regard to the two evaluation matrices. It clearly demonstrates that our \modelshort outperforms all baselines reported in PredRNN++ \cite{pmlr-v80-wang18b}. Specifically, \modelshort achieves the best SSIM of 0.811 and MSE of 85.41 per frame. Moreover, Fig. \ref{fig:mm} shows the training progress. From Fig.~\ref{sfig:mm-training}, we can see our proposed \modelshort converges much faster and smoother than other baselines. Also from prediction visualisations of Fig.~\ref{sfig:mm-example} we can see our proposed \modelshort has more accurate long term predictions. Specifically, at batch 20000, moving hand-written digits start to be recognisable at time step 41, whereas predictions of PredRNN++ are still blurry. This could be because with the curriculum of model $\mathcal{M}_1$, $\mathcal{M}_2$ can learn longer ranges as well as higher order dynamics, whereas in regular Seq2Seq, a later time step has to wait for the earlier time step to converge first due to the nature of RNN.

\section{Related Work}\label{sec:related}

\subsection{STSF with RNN}

Spatio-temporal sequence forecasting is a well studied research topic, and it is a part of time series prediction topic. In the literature, traditional machine learning based methods including Support Vector Machine 
(SVM) \cite{4840324}, Gaussian Process (GP)\cite{space-time-gp} and Auto-Regressive Integrated Moving Average (ARIMA) \cite{Box:1990:TSA:574978} have been proposed to deal with this problem. 

With recent advances in deep learning models in various domains, the research focus of spatio-temporal sequence forecasting has been redirected to deep models. For example, Shi \emph{et al.}~\cite{2018arXiv180806865S} proposed a Convolutional LSTM Recurrent Neural Network (RNN) for precipitation nowcasting. The main contribution of their work is that it models spatial correlations between two neighbouring time steps by replacing the linear transformation of LSTM with 2D-CNN,  which results in more compact spatial modelling and better performance. Zhang \emph{et al.} \cite{ZHANG2018147} proposed deep spatio-temporal residual networks to handle the citywide crowd flow prediction problem. In their work, temporal closeness, period and trends properties of crowded traffic, as well as external factors such as weather and events are considered and modelled by a fusion network. Similarly, Chen \emph{et al.} \cite{8594916} employed a 3D-CNN network to approach the problem. All these studies model the spatio-temporal data at a certain timestamp as an image-like format, each measurement at a location is treated as a pixel of an image. Therefore, modeling series of spatio-temporal data is the same as modeling videos (a series of images with a regular time interval). Forecasting a spatio-temporal sequence is highly relevant to  the problem of video frame prediction. Wang \emph{et al.} proposed PredRNN  \cite{NIPS2017_6689} and its following work PredRNN++  \cite{pmlr-v80-wang18b} to solve the video frame prediction problem. Their work is based on Seq2Seq with a custom RNN cell called Casual LSTM and a GHU unit.

Modeling a spatio-temporal sequence as a video-like format requires a rich collection of data in a variety of coordinate locations, also coordinate locations have to be a grid like format to be compatible as in the city crowd flow prediction. However, not all datasets satisfy these constraints. Another approach is to model spatio-temporal data as a vector, where each measurement of a location is a scalar of a row or column. For example, Ghaderi \emph{et al.} \cite{ghaderi2017deepforecast} proposed a LSTM-based model to predict wind speed across 57 measurement locations. Wang \emph{et al.} \cite{abs/1812.09467} proposed a deep uncertainty Seq2Seq model to predict weather for 10 weather stations across Beijing City. Yi \emph{et al.} \cite{Yi:2018:DDF:3219819.3219822} proposed a DeepAir model which consists of a spatial transformation component and a deep distributed fusion network to predict air quality. In another work, Liang \emph{et al.} \cite{ijcai2018-476}  proposed a multi-level attention Seq2Seq model called GeoMAN to model dynamics of spatio-temporal dependencies. All these studies above reveal spatio-temporal correlations are difficult to model, but are crucial for spatial-temporal predictions.

\subsection{Curriculum Learning Strategy}
%Our work is also inspired by the recent work in Generative Adversarial Network called PG-GAN \cite{karras2018progressive} and its following work FutureGAN \cite{sandrafuturegan}. PG-GAN introduced a progressive growing idea to start training with a small resolution of images. FutureGAN tried to adopt the progressive growing idea to video sequence prediction. Whereas, our work tries to adopt the progressive growing idea from spatial transformation to temporal modeling. 

Venktrman \emph{et al.} \cite{Venkatraman:2015:IMP:2888116.2888137} proposed a Data  As Demonstrator (DAD) model to improve multi-step prediction by feeding paired ground-truth and predicted word to the next step. The gap between single step prediction error and multi-step error in their work is similar to the gap between training and inference in our work. Bendigo \emph{et al.} \cite{Bengio:2015:SSS:2969239.2969370} further improve the idea by a Curriculum Learning \cite{10.1145/1553374.1553380} based approach to close the gap between training and inference by sampling previous ground-truth and previously predicted context by a changing probability. 

%Our work is also related to Mujika's recent work called Fast-Slow Recurrent Neural Networks  \cite{NIPS2017_7173}. In their work, they proposed a RNN architecture composed of multiscale RNNs that learns complex transition functions of different time scales. The fundamental difference is that they are stacking multiple RNN layers of different timescales while networks are fixed during training. Whereas in our approach, we first train an initial model at a larger timescale, then use the initial model to supervise a second model of smaller timescale. During training, the first model is progressively replaced, therefore our approach processes different timescales at a higher level and our approach is dynamic instead of statically fixed.

%\cite{karras2018progressive} PG-GAN
%
% \cite{sandrafuturegan} FutureGAN
 
% Fast-Slow Recurrent Neural Networks

\section{Conclusions and Future Work} \label{sec:conclusion}
In this paper, we propose \modelshort Sampling to bridge the gap between training and inference for Seq2Seq based STSF systems. By transforming the training process from a fully-supervised manner which utilises all available previous ground-truth to a less-supervised manner which replaces the ground-truth contexts with generated predictions. We also sample the target sequence from midway outputs from the intermediate model trained with bigger timescales with a carefully designed decaying strategy. Experiments on two datasets demonstrate that our proposed method achieves superior performance compared to all baseline methods, and show fast convergence speed as well as better long term accuracies. Two different types of datasets used in experiments prove the novelty and applicability of our proposed method.

Future studies are in two folds. First, an extensive ablation study along with a hyperparameter optimization study is required to further validate the applicability and utility of \model. Second, we can extend the progressive idea to spatial and temporal dimensions simultaneously for spatio-temporal sequence predictive learning.
\bibliographystyle{ecai}
%\fontsize{9.5pt}{10.5pt} \selectfont
\bibliography{bibliography}

\end{document}